%% file: main.tex
\documentclass{article} 
\usepackage{iclr2020_conference,times}

\input{math_commands.tex}

\usepackage[utf8]{inputenc} 
\usepackage[T1]{fontenc}    
\usepackage{hyperref}       
\usepackage{url}            
\usepackage{booktabs}       
\usepackage{amsfonts}       
\usepackage{nicefrac}       
\usepackage{microtype}      
\usepackage{amssymb,amsmath,amsthm,amsfonts}

\input preamble

\input aa_math

\title{\centering Solving Inverse Problems \\ by Joint Posterior Maximization \\ with a VAE Prior}

\author{Mario \textsc{González}\\
DMEL, CenUR LN\\
Universidad de la República\\
Salto, Uruguay \\
\texttt{mgonzalez@unorte.edu.uy} \\
\And
Andrés \textsc{Almansa} \\
MAP5 - CNRS \& Université de Paris \\
Paris, France\\
\texttt{andres.almansa@parisdescartes.fr} \\
\And
Mauricio \textsc{Delbracio} \& Pablo \textsc{Musé} \\
IIE - Facultad de Ingeniería\\
Universidad de la República \\
Montevideo, Uruguay \\
\texttt{\{mdelbra,pmuse\}@fing.edu.uy} \\
\And
Pauline \textsc{Tan}\\
LJLL - Sorbonne Université\\
Paris, France\\
\texttt{\small pauline.tan@sorbonne-universite.fr}
}

\iclrfinalcopy 
\begin{document}

\maketitle

\begin{abstract}
In this paper we address the problem of solving ill-posed inverse problems in imaging where the prior is a neural generative model.
Specifically we consider the decoupled case where the prior is trained once and can be reused for many different log-concave degradation models without retraining. Whereas previous \MAP-based approaches to this problem lead to highly non-convex optimization algorithms, our approach computes the joint (space-latent) \MAP\ that naturally leads to alternate optimization algorithms and to the use of a stochastic encoder to accelerate computations.
The resulting technique is called JPMAP because it performs Joint Posterior Maximization using an Autoencoding Prior.
We show theoretical and experimental evidence that the proposed objective function is quite close to bi-convex. Indeed it satisfies a weak bi-convexity property which is sufficient to guarantee that our optimization scheme converges to a stationary point.
 Experimental results also show the higher quality of the solutions obtained by our JPMAP approach with respect to other non-convex \MAP\ approaches which more often get stuck in spurious local optima.
\end{abstract}

\newpage

\tableofcontents

\newpage
\input body

\section*{Acknowledgements}
This work was partially funded by ECOS-Sud Project U17E04, by the French-Uruguayan Institute of Mathematics and interactions (IFUMI), by CSIC I+D (Uruguay) and by ANII (Uruguay) under Grant FCE\_1\_2017\_1\_135458. The authors would like to thank Warith Harchaoui, Alasdair Newson and Said Ladjal for useful discussions.


\appendix


\newpage
\input newdemo

\newpage



\bibliography{main}
\bibliographystyle{iclr2020_conference}

\end{document}

%% file: math_commands.tex
\usepackage{amsmath,amsfonts,bm}

\def\1{\bm{1}}

\def\vmu{{\bm{\mu}}}

\def\ve{{\bm{e}}}

\def\vu{{\bm{u}}}

\def\vx{{\bm{x}}}
\def\vy{{\bm{y}}}
\def\vz{{\bm{z}}}

\def\mA{{\bm{A}}}

\def\mSigma{{\bm{\Sigma}}}

\DeclareMathAlphabet{\mathsfit}{\encodingdefault}{\sfdefault}{m}{sl}
\SetMathAlphabet{\mathsfit}{bold}{\encodingdefault}{\sfdefault}{bx}{n}

\newcommand{\R}{\mathbb{R}}

\DeclareMathOperator*{\argmax}{arg\,max}
\DeclareMathOperator*{\argmin}{arg\,min}

%% file: preamble.tex
\usepackage{graphicx}
\usepackage{subfigure}

\usepackage{algorithm}
\usepackage{algorithmic}

\usepackage{amsthm}
\theoremstyle{plain}\newtheorem{theorem}{Theorem}
\theoremstyle{plain}\newtheorem{property}{Property}
\theoremstyle{plain}\newtheorem{proposition}{Proposition}
\theoremstyle{plain}\newtheorem{lemma}{Lemma}
\theoremstyle{plain}
\theoremstyle{plain}\newtheorem{assumption}{Assumption}
\theoremstyle{definition}

\usepackage{color}

\newcommand{\textcite}{\citet}
\newcommand{\parencite}{\citep}
\renewcommand{\cite}{\citep}

\newcommand{\MAP}{\textsc{map}}
\newcommand{\x}{\ensuremath{\vx}} 
\newcommand{\X}{X} 
\newcommand{\Y}{Y} 
\newcommand{\z}{\ensuremath{\vz}} 

\newcommand{\Z}{Z} 

\newcommand{\encoder}{\ensuremath{\mathsf{E}}}
\newcommand{\decoder}{\ensuremath{\mathsf{G}}}
\newcommand{\generator}{\decoder}
\newcommand{\encoderParams}{\ensuremath{\phi}}
\newcommand{\decoderParams}{\ensuremath{\theta}}
\newcommand{\muEncoder}{\vmu_{\encoderParams}}
\newcommand{\muDecoder}{\vmu_{\decoderParams}}
\newcommand{\SigmaEncoder}{\mSigma_{\encoderParams}}
\newcommand{\SigmaDecoder}{\mSigma_{\decoderParams}}

\newcommand{\Fdata}{\ensuremath{F}}
\newcommand{\Gprior}{\ensuremath{G}}
\newcommand{\Hcoupling}{\ensuremath{H}}
\newcommand{\Htheta}{\ensuremath{\Hcoupling_{\decoderParams}}}
\newcommand{\Kcoupling}{\ensuremath{K}}
\newcommand{\Kphi}{\ensuremath{\Kcoupling_{\encoderParams}}}

\newcommand{\pDecoder}{p_\decoderParams}
\newcommand{\ptheta}{\pDecoder}
\newcommand{\qEncoder}{q_\encoderParams}
\newcommand{\qphi}{\qEncoder}
\newcommand{\Normal}{\mathcal{N}} 


%% file: aa_math.tex
\newcommand{\NN}{\mathbb{N}} 
\newcommand{\RR}{\mathbb{R}} 

\newcommand{\pdf}{p} 

\newcommand{\PDF}[2]{\pdf_{#1} \left({#2}\right)} 

\newcommand{\ConditionalPDF}[4]{\pdf_{{#1} \vert {#2}} \left( {#3} \,\middle\vert\, {#4} \right)} 

\DeclareMathOperator*{\gd}{grad\,descent}
\newcommand{\GD}[3]{\gd_{#1} {#2},\,\text{starting from $#1 = #3$}}

\newcommand{\prox}{\operatorname{prox}}

%% file: body.tex
\section{Introduction and Related Work}

General inverse problems in imaging consist in estimating a clean image $\vx\in\R^n$ from noisy, degraded measurements $\vy\in\R^m$. In many cases the degradation model is known and its conditional density
$$ \ConditionalPDF{\Y}{\X}{\vy}{\vx} \propto e^{-\Fdata(\vx,\vy)} $$
is log-concave with respect to $\vx$. To illustrate this, let us consider the case where the negative log-conditional is quadratic with respect to $\vx$ 
\begin{equation} 
\label{eq:data-term}
\Fdata(\vx,\vy) = \frac{1}{2\sigma^2} \| \mA \vx - \vy \|^2. 
\end{equation}
This boils down to a linear degradation model that takes into account degradations such as, white Gaussian noise, blur, and missing pixels. When the degradation operator $\mA$ is non-invertible or ill-conditioned, or when the noise level $\sigma$ is high, obtaining a good estimate of $\vx$ requires prior knowledge on the image, given by $\PDF{X}{\vx} \propto e^{-\Gprior(\vx)}$. Variational and Bayesian methods in imaging are extensively used to derive MMSE or MAP estimators, 
\begin{equation} \label{eq:MAP}
\hat{\vx}_{\textsc{map}} = \argmax_\vx \ConditionalPDF{\X}{\Y}{\vx}{\vy} = \argmin_\vx \left \{\Fdata(\vx,\vy) + \Gprior(\vx)\right \}
\end{equation}
based on explicit priors like total variation \cite{Rudin1992,Chambolle04,Louchet2013,Pereyra2016}, or learning-based priors like patch-based Gaussian mixture models \cite{Zoran2011,Teodoro2018scene}.
\paragraph{Neural network regression.}
Since neural networks (NN) showed their superiority in image classification tasks~\cite{Krizhevsky2012} researchers started to look for ways to use this tool to solve inverse problems too. The most straightforward attempts employed neural networks as regressors to learn a risk minimizing mapping $\vy \mapsto \vx$ from many examples $(\vx_i,\vy_i)$ either
agnostically~\cite{dong2014learning,zhang2017beyond,zhang2018ffdnet, gharbi2016deep,schwartz2018deepisp,gao2019dynamic} 
or including the degradation model in the network architecture via unrolled optimization techniques \cite{gregor2010learning,Chen2017,diamond2017unrolled,gilton2019neumann}.
\paragraph{Implicitly decoupled priors.}
The main drawback of neural networks regression is that they require to retrain the neural network each time a single parameter of the degradation model changes.
To avoid the need for retraining, another family of approaches seek to \emph{decouple} the NN-based learned image prior from the degradation model.
A popular approach within this methodology are \emph{plug \& play} methods. Instead of directly learning the log-prior $-\log \PDF{X}{\vx} = \Gprior(\vx) + C$, these methods seek to learn an approximation of its gradient $\nabla\Gprior$ \cite{Bigdeli2017,Bigdeli2017a} or proximal operator $\prox_\Gprior$ \cite{meinhardt2017learning,Zhang2017,chan2017plug,ryu2019plug}, by replacing it by a denoising NN. Then, these approximations are used in an iterative optimization algorithm to find the corresponding MAP estimator in equation~\eqref{eq:MAP}. 
\paragraph{Explicitly decoupled priors.}
Plug \& play approaches became very popular because of their convenience but obtaining convergence guarantees under realistic conditions is quite challenging. Indeed, the actual prior is unknown, and the existence of a density whose proximal operator is well approximated by a neural denoiser is most often not guaranteed \citep{reehorst2018regularization}, unless the denoiser is retrained with specific constraints \citep{ryu2019plug,gupta2018cnn,shah2018solving}. In our experience these inconsistencies may result in sub-optimal solutions that introduce undesirable artifacts. 
It is tempting to use neural networks to learn an explicit prior for images. For instance one could use a generative adversarial network (GAN) to learn a generative model for $X=\generator(\Z)$ with $\Z\sim N(0,I)$ a latent variable. Nevertheless, current attempts \cite{bora2017compressed} to use such a generative model as a prior to estimate $\hat{\vx}_{\textsc{map}}$ in \eqref{eq:MAP} lead to a highly non-convex optimization problem. Indeed, the posterior writes
\begin{align*}
    \ConditionalPDF{\X}{\Y}{\vx}{\vy} & = 
    \int \ConditionalPDF{\Y}{\X}{\vy}{\vx} \ConditionalPDF{\X}{\Z}{\vx}{\vz}  \PDF{\Z}{\vz} d\vz\\
    & = \ConditionalPDF{\Y}{\X}{\vy}{\vx} \int \ConditionalPDF{\X}{\Z}{\vx}{\vz}  \PDF{\Z}{\vz} d\vz \\
    & = \ConditionalPDF{\Y}{\X}{\vy}{\vx} \PDF{\Z}{\generator^{-1}(\vx)},
\end{align*}
where the last equality follows from $\ConditionalPDF{\X}{\Z}{\vx}{\vz} = \delta(x-G(z))$. Therefore, 
\begin{equation}\label{eq:Bora}
\begin{split}
\hat{\vx}_{\textsc{map}} & = \argmax_x \ConditionalPDF{\X}{\Y}{\vx}{\vy} \\
& = \generator\left(\argmax_z \left \{\ConditionalPDF{\Y}{\X}{\vy}{\generator(\vz)}\PDF{\Z}{\vz} \right \}\right) \\
& = \generator\left(\argmin_z \left \{\Fdata(\generator(\vz),\vy) + \frac{1}{2}\|\vz\|^2\right \}\right ).
\end{split}
\end{equation}
Convergence guarantees for this problem are of course extremely difficult to establish, and our experimental results in Section~\ref{sec:experiments} on the CSGM approach by \citet{bora2017compressed} confirm this.

A possible workaround to avoid minimization over $\vz$ could consist in learning an encoder network $\encoder$ (inverse of \generator) to directly minimize over $\vx$. This does not help either because an intractable term appears when we develop
$$\PDF{\X}{\vx} = \PDF{Z}{\encoder(\vx)} \det\left(\left(\frac{\partial \encoder}{\partial \vx}\right)^T\left(\frac{\partial \encoder}{\partial \vx}\right)\right)^{1/2} \delta_{\generator(\encoder(\vx))=\vx}(\vx) $$ 
via the push-forward measure.

\paragraph{Proposed method: Joint $\textsc{map}_{x,z}$.} In order to overcome the limitations of the previous approach, in this work we show that the numerical solution of the explicitly decoupled approach is greatly simplified when we introduce two modifications:
\begin{itemize}
    \item Given the noisy, degraded observation $y$, we maximize the joint posterior density $\ConditionalPDF{\X,\Z}{\Y}{\vx,\vz}{\vy}$ instead of the usual posterior $\ConditionalPDF{\X}{\Y}{\vx}{\vy}$;
    \item We use both a (deterministic or stochastic) generator and a stochastic encoder.
\end{itemize}
In addition, we show that for a particular choice of the stochastic decoder the maximization of the joint log-posterior becomes a bi-concave optimization problem or approximately so. And in that case, an extension of standard bi-convex optimization results~\cite{Gorski2007} show that the algorithm converges to a stationary point that is a partially global optimum.

The remainder of this paper is organized as follows. In Section~\ref{sec:JPMAP_framework} we derive a model for the joint conditional posterior distribution of space and latent variables $\vx$ and $\vz$, given the observation $\vy$. This model makes use of a generative model, more precisely a VAE with Gaussian decoder. We then propose an alternate optimization scheme to maximize for the joint posterior model, and state convergence guarantees. Section~\ref{sec:experiments} presents first a set of experiments that illustrates the convergence properties of the optimization scheme. We then test our approach on classical image inverse problems, and compare its performance with state-of-the-art methods. Concluding remarks are presented in Section~\ref{sect:future_work}.

\section{Joint Posterior Maximization with Autoencoding Prior (JPMAP)}
\label{sec:JPMAP_framework}

\subsection{Variational Autoencoders as Image Priors}
\label{subsec:VAEs_priors}

In this work we construct an image prior based on a variant of the variational autoencoder~\cite{Kingma2014} (VAE).
Like GANs and other generative models, VAEs allow to obtain samples from an unknown distribution $p_X$ by taking samples of a latent variable $\Z$ with known distribution $\mathcal{N}(0,I)$, and feeding these samples through a learned generator network.  
For VAEs the generator (or decoder) network with parameters \decoderParams \, can be deterministic or stochastic and it learns
$$\ConditionalPDF{\X}{\Z}{\vx}{\vz} = \ptheta(\vx|\vz),$$
whereas the stochastic encoder network with parameters \encoderParams, approximates
$$\ConditionalPDF{\Z}{\X}{\vz}{\vx} \approx \qphi(\vz|\vx).$$

Given a VAE we could plug in the approximate prior
\begin{equation}
\label{eq:approx-prior}
\PDF{\X}{\vx} = \frac{\ptheta(\vx|\vz)\, \PDF{\Z}{\vz}}{\ConditionalPDF{\Z}{\X}{\vz}{\vx}} 
          \approx \frac{\ptheta(\vx|\vz)\, \PDF{\Z}{\vz}}{\qphi(\vz|\vx)} 
\end{equation}
in~\eqref{eq:MAP} to obtain the corresponding MAP estimator, but this leads to a numerically difficult problem to solve. Instead, we propose to maximize the joint posterior $\ConditionalPDF{\X,\Z}{\Y}{\vx,\vz}{\vy}$ over $(\vx,\vz)$ which is equivalent to minimize
\begin{align}\label{eq:JPMAP}
        J_1(\vx,\vz) := -\log \ConditionalPDF{\X,\Z}{\Y}{\vx,\vz}{\vy} &= 
        -\log \ConditionalPDF{\Y}{\X,\Z}{\vy}{\vx,\vz} \ptheta(\vx\,|\,\vz)\PDF{\Z}{\vz} \\
        &= \Fdata(\vx,\vy) + \Htheta(\vx,\vz) + \frac{1}{2}\|\vz\|^2.
\end{align}
Note that the first term is quadratic in $\vx$ (assuming~\eqref{eq:data-term}), the third term is quadratic in $\vz$ and all the difficulty lies in the coupling term $\Htheta(\vx,\vz)=-\log \ptheta(\x\,|\,\z)$. 
For Gaussian decoders~\cite{Kingma2014}, the latter can be written as
\begin{equation}
 \Htheta(\vx,\vz)= 
    \frac{1}{2}\left(n\log(2\pi) + \log\det\Sigma_\theta(\vz) + \|\SigmaDecoder^{-1/2}(\vz)(\vx-\muDecoder(\vz))\|^2\right).
\end{equation}
which is also convex in $\vx$. Hence, minimization with respect to $\vx$ takes the convenient closed form:
\begin{equation}\label{eq:xmin}
    \argmin_\vx J_1(\vx,\vz) = 
    \left(\mA^T\mA + \sigma^2\SigmaDecoder^{-1}(\vz)\right)^{-1}\left(\mA^T\vy + \sigma^2\SigmaDecoder^{-1}(\vz)\muDecoder(\vz)\right).
\end{equation}

Unfortunately the coupling term $\Hcoupling$ and hence $J_1$ is a priori non-convex in $\vz$. 
As a consequence the $\vz$-minimization problem
\begin{equation}\label{eq:zmin-exact}
    \argmin_\vz J_1(\vx,\vz) 
 \end{equation}
 is a priori more difficult.
However, for Gaussian encoders, VAEs provide an approximate expression for this coupling term which is quadratic in $\vz$. Indeed, given the equivalence  
$$
\ptheta(\vx\,|\,\vz) \, \PDF{\Z}{\vz}
= \PDF{\X,\Z}{\vx,\vz}
= \ConditionalPDF{ \Z }{\X}{\vz}{\vx} \, \PDF{\X}{\vx}
\approx \qphi(\vz\,|\,\vx) \, \PDF{\X}{\vx}$$
we have that 
\begin{equation}\label{eq:autoencoder-approx}
    \Htheta(\vx,\vz) + \frac{1}{2}\|\vz\|^2 \approx \Kphi(\vx,\vz) - \log \PDF{\X}{\vx}.
\end{equation}
where $\Kphi(\vx,\vz)=-\log \qphi(\vz\,|\,\vx)$. Therefore, this new coupling term becomes
\begin{align}
 \Kphi(\vx,\vz) & = -\log \mathcal{N}\left(\vz; \muEncoder(\vx),\SigmaEncoder(\vx)\right) \\
              & = \frac{1}{2}\left(k\log(2\pi) + \log \det \SigmaEncoder(\vx) + \|\SigmaEncoder^{-1/2}(\vx)(\vz-\muEncoder(\vx))\|^2\right),
\end{align}
which is quadratic in $\vz$. This provides an approximate expression for the energy~\eqref{eq:JPMAP} that we want to minimize, namely 
\begin{equation}\label{eq:JPMAPapprox}
        J_2(\vx,\vz) := \Fdata(\vx,\vy) + \Kphi(\vx,\vz) - \log \PDF{\X}{\vx}
        \approx J_1(\vx,\vz).
\end{equation}
This approximate functional is quadratic in $\vz$, and minimization with respect to this variable yields
\begin{equation}\label{eq:zmin}
    \argmin_\vz J_2(\vx,\vz) =
    \muEncoder(\vx).
\end{equation}

\subsection{Alternate Joint Posterior Maximization}
\label{sect:assumptions}

The previous observations suggest to adopt alternate scheme to minimize $-\log \ConditionalPDF{\X,\Z}{\Y}{\vx,\vz}{\vy}$ in order to solve the inverse problem. We begin our presentation by assuming that the approximation of $J_1$ by $J_2$ is exact; then we propose an adaptation for the non-exact case and we explore its convergence properties.

To begin with we shall consider the following (strong) assumption:
\begin{assumption}\label{exact-approximation}
The approximation in~\eqref{eq:JPMAPapprox} is exact, \emph{i.e.} $J_1=J_2$.
\end{assumption}

Under this assumption, the objective function is biconvex and alternate minimization takes the simple and fast form depicted in Algorithm~\ref{alg:JPMAPexact}, which can be shown to converge to a partial optimum, as stated in Proposition~\ref{thm:convergence-exact} below. Note that the minimization in step~2 of Algorithm~\ref{alg:JPMAPexact} does not require the knowledge of the unknown term $- \log \PDF{\X}{\vx}$ in Equation~\eqref{eq:JPMAPapprox} since it does not depend on $\vz$.

\renewcommand{\algorithmiccomment}[1]{\hfill // #1}
\begin{algorithm}
\caption{Joint posterior maximization - exact case}
\label{alg:JPMAPexact}
\begin{algorithmic}[1]
\REQUIRE Measurements $\vy$, Autoencoder parameters \decoderParams, \encoderParams, Initial condition $\vx_0$
\ENSURE $\hat{\vx}, \hat{\vz} = \argmax_{\vx,\vz} \ConditionalPDF{\X,\Z}{\Y}{\vx,\vz}{\vy}$

\FOR{$n:=0$ \TO maxiter}
\STATE $\vz_{n+1} := \argmin_\vz J_2(\vx_n,\vz)$ \COMMENT{Quadratic problem in~\eqref{eq:xmin}}
\STATE $\vx_{n+1} := \argmin_\vx J_1(\vx,\vz_{n+1})$ \COMMENT{Quadratic problem in~\eqref{eq:zmin}} 
\ENDFOR 
\RETURN $\vx_{n+1}, \vz_{n+1}$

\end{algorithmic}
\end{algorithm}

\begin{proposition}[Convergence of Algorithm~\ref{alg:JPMAPexact}]\label{thm:convergence-exact}
Under Assumption~\ref{exact-approximation} we have that :
\begin{enumerate}
\item The sequence $\left\lbrace J_1(\vx_{n},\vz_{n}) \right\rbrace$ generated by Algorithm~\ref{alg:JPMAPexact} converges
monotonically when $n \to \infty$.
The sequence $\left\lbrace (\vx_{n},\vz_{n})  \right\rbrace$ has at least one accumulation point.

\item All accumulation points are partial optima of $J_1$ and they all have the same function value.
\end{enumerate}
If in addition $J_1$ is differentiable then:
\begin{enumerate}\setcounter{enumi}{2}
\item The set of all accumulation points are stationary points of $J_1$ and they form a connected, compact set.
\end{enumerate}
\end{proposition}

The proof of this proposition is given in Appendix~\ref{sec:proofs}, and follows the same arguments as in \citep{Gorski2007,Aguerrebere2014b}.
Note that the third part requires that $J_1$ be differentiable, which is the case if we use a differentiable activation function like the Exponential Linear Unit (ELU)~\citep{Clevert2016} with $\alpha=1$, instead of the more common ReLU activation function.

When the autoencoder approximation in~\eqref{eq:JPMAPapprox} is not exact (Assumption~\ref{exact-approximation}), the algorithm needs some additional steps to ensure that the energy we want to minimize, namely $J_1$, actually decreases.
Nevertheless, the approximation provided by $J_2$ is still very useful since it provides  a fast and accurate heuristic to initialize the minimization of $J_1$. This method is presented in Algorithm~\ref{alg:JPMAP2}.

\renewcommand{\algorithmiccomment}[1]{\hfill // #1}
\begin{algorithm}
\caption{Joint posterior maximization - approximate case}
\label{alg:JPMAP2}
\begin{algorithmic}[1]
\REQUIRE Measurements $\vy$, Autoencoder parameters \decoderParams, \encoderParams, Initial conditions $\vx_0, \vz_0$ 
\ENSURE $\hat{\vx}, \hat{\vz} = \argmax_{\vx,\vz} \ConditionalPDF{\X,\Z}{\Y}{\vx,\vz}{\vy}$

\FOR{$n:=0$ \TO maxiter}
\STATE $\vz^{0} := \GD{\vz}{ J_1(\vx_n,\vz)}{\vz_{n}}$
\STATE $\vz^{1} := \argmin_\vz J_2(\vx_n,\vz)$ \COMMENT{Quadratic problem}
\STATE $\vz^{2} := \GD{\vz}{J_1(\vx_n,\vz)}{\vz^{1}}$
\STATE $\vz_{n+1} := \argmin_{\vz \in \{\vz^0,\vz^1,\vz^2\}} J_1(\vx_n,\vz) $
\STATE $\vx_{n+1} := \argmin_\vx J_1(\vx,\vz_{n+1})$ \COMMENT{Quadratic problem} 
\ENDFOR 
\RETURN $\vx_{n+1}, \vz_{n+1}$

\end{algorithmic}
\end{algorithm}

Algorithm~\ref{alg:JPMAP2} provides also a useful tool for diagnostics. Indeed, the comparison of the evaluation of $J_1(\vx_n,\vz)$ in $\vz^{0}, \vz^{1}, \vz^{2}$ performed in step 5 permits to assess the evolution of the approximation of $J_1$ by $J_2$. 

Our experiments with Algorithm~\ref{alg:JPMAP2} (Section~\ref{sec:qualitative-JPMAP2}) show that during the first few iterations (where the approximation provided by $J_2$ is good enough) $\vz^2$ reaches convergence faster than $\vz^0$. After a critical number of iterations the opposite is true (the initialization provided by the previous iteration is better than the $J_2$ approximation, and $\vz^0$ converges faster).

These observations suggest that a faster execution, with the same convergence properties, can be achieved by the variant in Algorithm~\ref{alg:JPMAP3}. 

The fastest alternative is equivalent to Algorithm~\ref{alg:JPMAPexact} as long as the approximate energy minimization decreases the actual energy. When this is not the case it will take a slower route similar to Algorithm~\ref{alg:JPMAP2}.

\renewcommand{\algorithmiccomment}[1]{\hfill // #1}
\begin{algorithm}
\caption{Joint posterior maximization - approximate case (faster version)}
\label{alg:JPMAP3}
\begin{algorithmic}[1]
\REQUIRE Measurements $\vy$, Autoencoder parameters \decoderParams, \encoderParams, Initial conditions $\vx_0, \vz_0$, iterations $n_{\min} < n_{\max}$
\ENSURE $\hat{\vx}, \hat{\vz} = \argmax_{\vx,\vz} \ConditionalPDF{\X,\Z}{\Y}{\vx,\vz}{\vy}$

\FOR{$n:=0$ \TO $n_{\max}$}
\STATE $\vz^{1} := \argmin_\vz J_2(\vx_n,\vz)$
\COMMENT{Quadratic problem}
\STATE $\vz^{0} : = \vz_{n}$
\STATE $\vz^* : = \argmin_{\vz\in\lbrace\vz^0,\vz^1\rbrace} J_1(\vx_n,\vz)$
\IF{$J_1(\vx_n,\vz^{1}) > J_1(\vx_n,\vz_n)$ \OR $n>n_{\min}$} 
    \STATE $\vz_{n+1} := \GD{\vz}{J_1(\vx_n,\vz)}{\vz^{*}}$
\ELSE
    \STATE $\vz_{n+1} := \vz^{1}$ \COMMENT{Faster alternative while $J_2$ is good enough}
\ENDIF
\STATE $\vx_{n+1} := \argmin_\vx J_1(\vx,\vz_{n+1})$ \COMMENT{Quadratic problem} 
\ENDFOR 
\RETURN $\vx_{n+1}, \vz_{n+1}$

\end{algorithmic}
\end{algorithm}

Algorithm~\ref{alg:JPMAP2} is still quite fast when $J_2$ provides a sufficiently good approximation. Even if we cannot give a precise definition of what {\em sufficiently good} means, the sample comparison of $\Kphi$ and $\Htheta$ as functions of $\vz$, displayed in Figure~\ref{fig:encoder_approximation}, shows that the approximation is fair enough in the sense that it preserves the global structure of $J_2$. The same behavior was observed for a large number of random tests. In particular, these simulations show that for every tested $\vx$, the function $\vz \mapsto J_1(\vx,\vz)$ exhibits a unique global minimizer. This justifies the following assumption (which is nevertheless much weaker than the previous Assumption~\ref{exact-approximation}):

\begin{assumption}\label{z-convexity1} \hfill \break
{\em(A)} The z-minimization algorithm $\gd_\z J_1(\x,\z)$ converges to a global minimizer of $\z \mapsto J_1(\x,\z)$, when initialized at $\vz^1 = \argmin_\z J_2(\x,\z)$ or at any $\vz$ such that $J_1(\vx,\vz)\leq J_1(\vx,\vz^1)$. 

{\em(B)} The map $\z \mapsto J_1(\x,\z)$ has a single global minimizer.
\end{assumption}

Under this assumption we have the following result for Algorithm~\ref{alg:JPMAP2}:

\begin{proposition}[Convergence of Algorithms~\ref{alg:JPMAP2} and \ref{alg:JPMAP3}]\label{thm:convergence-approx}
Under Assumption 2A we have that:
\begin{enumerate}
\item The sequence $\left\lbrace J_1(\vx_{n},\vz_{n}) \right\rbrace$ generated by Algorithms~\ref{alg:JPMAP2} and \ref{alg:JPMAP3} converges monotonically when $n \to \infty$\\
The sequence $\left\lbrace (\vx_{n},\vz_{n})  \right\rbrace$ has at least one accumulation point.
\item All accumulation points are partial optima of $J_1$ and they all have the same function value.
\end{enumerate}
If in addition $J_1$ is continuously differentiable and Assumption 2B holds, then:
\begin{enumerate}\setcounter{enumi}{2}
\item All accumulation points are stationary points of $J_1$.
\end{enumerate}
\end{proposition}

The proof of this proposition is detailed in the appendix. The first two parts are similar to the proof of Proposition~\ref{thm:convergence-exact}, but the last part uses a different argument. Indeed we cannot use \citep[Thm 4.9]{Gorski2007}, because we do not assume here that $J_1$ is bi-convex.

\section{Experimental results}
\label{sec:experiments}
\subsection{AutoEncoder and dataset}
\label{subsec:exp_setup}
In order to test our joint prior maximization model we first train a Variational Autoencoder like in \citep{Kingma2014} on the training data of MNIST handwritten digits \citep{MNIST}.

The \emph{stochastic encoder} takes as input an image $\vx$ of $24\times24 = 784$ pixels and produces as an output the mean and (diagonal) covariance matrix of the Gaussian distribution $q_\phi(\vz|\vx)$, where the latent variable $\vz$ has dimension $12$.
The architecture of the encoder is composed of 4 fully connected layers with ELU activations (to preserve continuous differentiability). The sizes of the layers are as follows:
$$784\to 500\to 500 \to (12 + 12).$$
Note that the output is of size $12+12$ in order to encode the mean and diagonal covariance matrix, both of size 12.

The \emph{stochastic decoder} takes as an input the latent variable $\vz$ and outputs the mean and covariance matrix of the Gaussian distribution $p_\theta(x|z)$. Following \citep{TwoStageVAE} we chose here an isotropic covariance $\SigmaDecoder(\vz)=\gamma I$ where $\gamma>0$ is trained, but independent of $\vz$. This choice simplifies the minimization problem (\ref{eq:zmin-exact}), because the term $\det\Sigma_\theta(\vz)$ (being constant) has no effect on the $\vz$-minimization.\\
The architecture of the decoder is also composed of 4 fully connected layers with ELU activations (to preserve continuous differentiability). The sizes of the layers are as follows:
$$12\to 500\to 500 \to 784.$$
Note that the covariance matrix is constant, so it does not augment the size of the output layer which is still $784=24\times24$ pixels.

We train this architecture using TensorFlow\footnote{Code reused from \url{https://github.com/daib13/TwoStageVAE} \cite{TwoStageVAE}} with batch size 64 and Adam algorithm for 400 epochs with learning rate 0.0001 (halving every 150 epochs) and rest of the parameters as default.

\begin{figure}
  \subfigure[Reconstruction examples]{
  \label{fig:VAE_reconstruction}
  \includegraphics[width=0.30\textwidth]{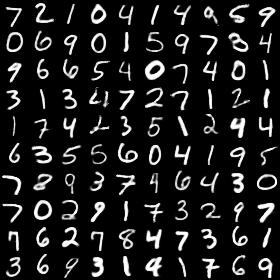}
  }
  \subfigure[Random samples]{
  \label{fig:VAE_samples}
  \includegraphics[width=0.30\textwidth]{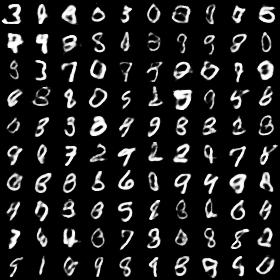}
  }
  \subfigure[encoder approximation]{
  \label{fig:encoder_approximation}
  \includegraphics[width=0.33\textwidth]{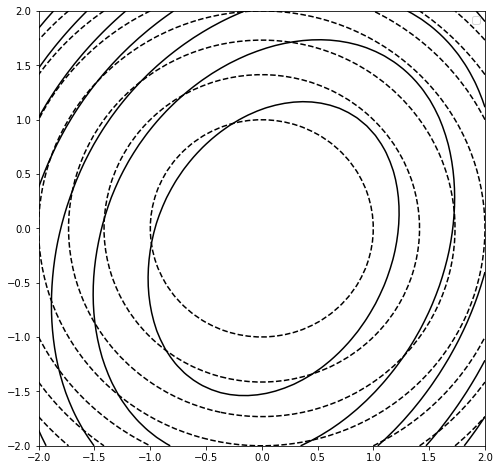}
  }
  \caption{\emph{Visualisation of the trained VAE on MNIST}.
  \emph{(a)} Reconstruction examples, \emph{i.e.} $\muDecoder(\muEncoder(\vx))$ for 100 samples of $\vx$ from the test set.
  \emph{(b)} Random samples, \emph{i.e.} $\muDecoder(\vz)$ for 100 random samples $\vz\sim\Normal(0,Id)$.
  \emph{(c)} 
  \emph{Encoder approximation:} Contour plots of $-\log p_\theta(\vx|\vz) +\frac{1}{2}\|\vz\|^2$ and $-\log q_\phi(\vz|\vx)$ for a fixed $\vx$ and for a random 2D subspace in the $\vz$ domain (the plot shows $\pm 2 \SigmaEncoder^{1/2}$ around $\muEncoder$). Observe the relatively small gap between the true posterior $p_\theta(\vz|\vx)$ and its variational approximation $q_\phi(\vz|\vx)$. This figure shows some evidence of partial $\z$-convexity of $J_1$ around the minimum of $J_2$, but it does not show how far is $\z^1$ from $\z^2$.
  }
  \label{fig:trained_VAE}
 \end{figure}

The subjective quality of the trained VAE is illustrated in Figure~\ref{fig:trained_VAE}, including reconstruction examples (Figure~\ref{fig:VAE_reconstruction}) and random samples (Figure~\ref{fig:VAE_samples}). Figure~\ref{fig:encoder_approximation} shows that the encoder approximation $q_\phi(\vz|\vx)$ of the true posterior $p_\theta(\vz|\vx)$ is not perfect but is quite tight. It also shows that the true posterior $p_\theta(\vz|\vx)$ is pretty close to log-concave near the maximum of $q_\phi(\vz|\vx)$.

\subsection{Empirical Validation of Assumption 2}
\label{sec:qualitative-JPMAP2}
\begin{figure}
  \subfigure[Energy evolution, initializing with $\mathcal{N}(0,I)$.]{
  \label{fig:assumption2A}
  \includegraphics[width=0.5\textwidth]{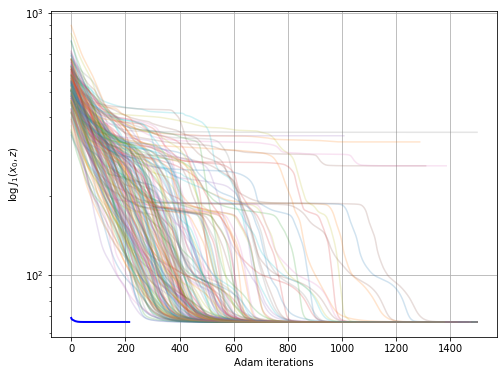}
  }
  \subfigure[Distance to the optimum at each iteration of (a).]{
  \label{fig:assumption2B}
  \includegraphics[width=0.5\textwidth]{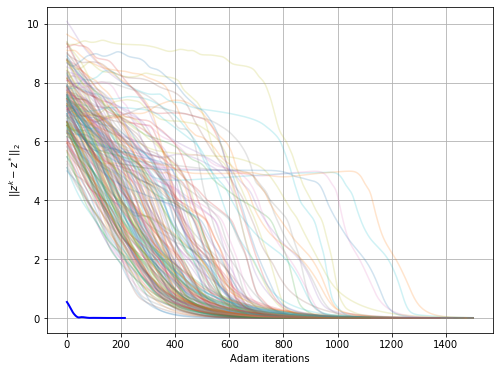}
  }
  \subfigure[Energy evolution, initializing with $q_\phi(z|x_0)$.]{
  \label{fig:assumption2A_b}
  \includegraphics[width=0.5\textwidth]{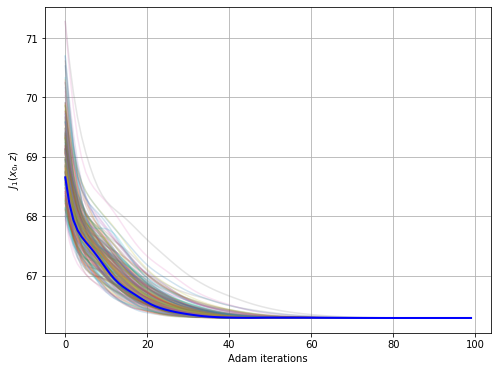}
  }
  \subfigure[Distance to the optimum at each iteration of (c).]{
  \label{fig:assumption2B_b}
  \includegraphics[width=0.5\textwidth]{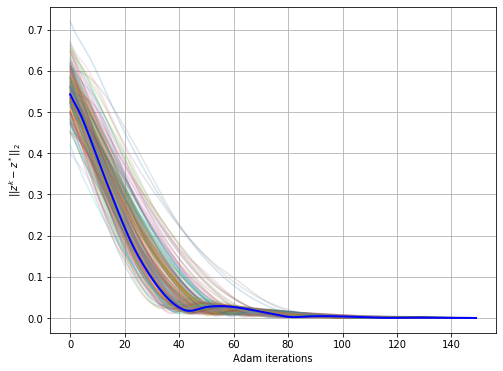}
  }
  \caption{
  \emph{Experimental validation of Assumption 2:} We take $x_0$ from the \emph{test} set of MNIST and minimize $J_1(\vx_0,\vz)$ with respect to $\vz$ using gradient descent from different initializations $\vz_0$. The blue thick curve represents the trajectory if we initialize at the encoder approximation $\vz^1=\argmin_\vz J_2(\vx_0,\vz)=\muEncoder(\vx_0)$. \emph{(a)} and \emph{(c)}: Plots of the energy iterates $J_1(\vx_0,\vz_k)$. \emph{(b)} and \emph{(d)}: $\ell^2$ distances of each trajectory with respect to the global optimum $z^*$. (a) and (b): Evolution of Adam on 200 random Gaussian initializations. (c) and (d): Initializing with random samples taken from the posterior approximation $q_\phi(\vz|\vx_0)$ given by the encoder. 
  \emph{Conclusion:} Observe that all initializations $\vz^0$ such that $J_1(\vx_0,\vz_0)\leq J_1(\vx_0,\vz^1)$ do converge to a unique global minimizer, and so do many other initializations.
  }
 \end{figure}

As we have shown in Section~\ref{sec:JPMAP_framework}, our simplest Algorithm~\ref{alg:JPMAPexact} is only guaranteed to converge when the encoder approximation is exact thus ensuring bi-convexity of $J_1=J_2$.

In practice an exact autoencoder approximation is difficult to achieve, and in particular the VAE trained in the previous section has a small gap between $J_1$ and $J_2$. To consider this case we proposed Algorithms~\ref{alg:JPMAP2} and \ref{alg:JPMAP3} which are guaranteed to converge under weaker quasi-bi-convex conditions stated in Assumption~\ref{z-convexity1}. 

In this section we experimentally check that the VAE we trained in the previous section actually verifies such conditions. We do so by selecting a random $\vx_0$ from MNIST test set and by computing $\vz^*(\vz_0) := \gd_\vz J_1(\vx_0,\vz)$ with different initial values $\vz_0$. These experiments were performed using the Adam minimization algorithm with learning rate equal to $0.01$. 

Figures~\ref{fig:assumption2A}~and~\ref{fig:assumption2A_b} show that $\vz^*(\vz_0)$ reaches the global optimum for initializations $\vz_0$ chosen under far less restrictive conditions than those required by Assumption~\ref{z-convexity1}(A).
Indeed from 200 random initializations $\vz_0 \sim \Normal(0,I)$, 195 reach the same global minimum, whereas 5 get stuck at a higher energy value. However these 5 initial values have energy values $J_1(\vx_0,\vz_0) \gg J_1(\vx_0,\vz^1)$ far larger than those of the encoder initialization $\vz^1 = \muEncoder(\vx_0)$, and are thus never chosen by Algorithms~\ref{alg:JPMAP2}~and~\ref{alg:JPMAP3}.

This experiment validates Assumption~\ref{z-convexity1}(A). In addition, it shows that we cannot assume $\vz$-convexity: The presence of plateaux in the trajectories of many random initializations as well as the fact that a few initializations do not lead to the global minimum indicates that $J_1$ may not be everywhere convex with respect to $\vz$. However it still satisfies the weaker  Assumption~\ref{z-convexity1}(A) which is sufficient to prove convergence in Theorem~\ref{thm:convergence-approx}.

In Figures~\ref{fig:assumption2B}~and~\ref{fig:assumption2B_b} we display the distances of each trajectory (except for the 5 outliers) to the global optimum $\vz^*$ (taken as the median over all initializations $\vz_0$ of the final iterates $\vz^*(\vz_0)$); note that this optimum is always reached, which proves that $\vz \mapsto J_1(\vx_0,\vz)$ has unique global minimizer.

\subsection{Image restoration experiments}

\emph{Choice of $\vx_0$:} In the previous section, our validation experiments used a random $\vx_0$ from the data set as initialization. When dealing with an image restoration problem, Algorithms~\ref{alg:JPMAP2}~and~\ref{alg:JPMAP3} require an initial value of $\vx_0$ to be chosen. In all experiments we choose this initial value as the simplest possible inversion algorithm, namely the regularized pseudo-inverse of the degradation matrix:
$$\vx_0 = A^\dag y = (A^TA+\varepsilon Id)^{-1}A^T y.$$ 
\emph{Choice of $n_{\min}$:} After a few runs of Algorithm~\ref{alg:JPMAP2} we find that in most cases, during the first two or three iterations $\vz^1$ decreases the energy with respect to the previous iteration. But after at most five iterations the autoencoder approximation is no longer good enough and we need to perform gradient descent on $\vz$ in order to further decrease the energy. Based on these findings we set $n_{\min} = 5$ in Algorithm~\ref{alg:JPMAP3} for all experiments.

\begin{figure}
  \subfigure[Results on inpainting.]{
  \label{fig:results_inpainting}
  \includegraphics[width=0.27\textwidth]{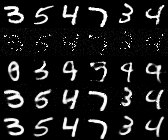}
  }
  \subfigure[Results on compressed sensing.]{
  \label{fig:results_compressed_sensing}
  \includegraphics[width=0.34\textwidth]{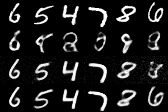}
  }
  \subfigure[MSE comparison with CSGM.]{
  \label{fig:mse_inpainting}
  \includegraphics[width=0.34\textwidth]{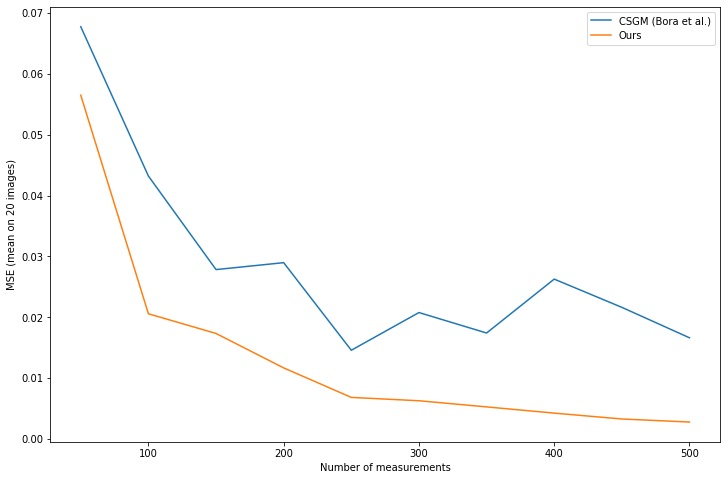}
  }
  \caption{\emph{Experimental results on MNIST.} Comparison with CSGM algorithm \cite{bora2017compressed}. 
  \emph{(a)} Some selected results from the inpainting experiment with $m=100$ 
  (12.8\%) known pixels.
  From top to bottom: original image $\vx^*$, corrupted image $\tilde \vx$, restored by \cite{bora2017compressed}, restored image $\hat \vx$ by our framework (with Gaussian decoder) and the reconstruction of the original image by the VAE $\muDecoder(\muEncoder(\vx^*))$ which can be seen as the best possible reconstruction if we use this model. 
  \emph{(b)} Same as (a) in a compressed sensing experiment with $m=100$ (12.8\%) random measurements (without showing the second row). 
  \emph{(c)} MSE mean of the reconstruction of 20 images for the compressed sensing experiment, varying the number of measurements $m$.
  \emph{Conclusion:} Our algorithm performs consistently better than CSGM. In addition our algorithm gets less often stuck in spurious local minima.}
  \label{fig:results_comparison}
 \end{figure}

Figure~\ref{fig:results_comparison} displays some selected results of compressed sensing and inpainting experiments on MNIST using the proposed approach. Figure~\ref{fig:results_inpainting} shows an \emph{inpainting} experiment with 80\% of missing pixels and Gaussian white noise with $\sigma=2/255$. Figure~\ref{fig:results_compressed_sensing} shows a \emph{compressed sensing} experiment with $m=100$ random measurements and Gaussian white noise with $\sigma=2/255$. For comparison we provide also the result of another decoupled approach proposed by \textcite{bora2017compressed} with $\lambda=0.1$ as suggested in the paper, which uses the same generative model to compute the MAP estimator as in Equation~\eqref{eq:MAP}, but does not make use of the encoder.\footnote{Since \textcite{bora2017compressed} does not provide code, we implemented our own version of their algorithm and utilize the same trained VAE as encoder $\phi$ for both algorithms.}

As we can see in Figure \ref{fig:results_comparison}, the proposed method significantly outperforms CSGM. There are still some failure cases (see last column in Figure~\ref{fig:results_compressed_sensing}). However, in the vast majority of cases our alternate minimization scheme does not get stuck in local optima, as CSGM does.


\section{Conclusions and Future work}\label{sect:future_work}

In this work we presented a new framework to solve convex inverse problems with priors learned in the latent space via variational autoencoders. Unlike similar approaches like CSGM \cite{bora2017compressed} which learns the prior based on generative models, our approach is based on a generalization of alternate convex search to quasi-biconvex functionals. This quasi-biconvexity is the result of considering the joint posterior distribution of latent and image spaces. As a result, the proposed approach provides stronger convergence guarantees. 
Experiments on inpainting and compressed sensing confirm this, since our approach gets stuck much less often in spurious local minima than CSGM, which is simply based on gradient descent of a highly non-convex functional. This leads to restored images which are significantly better in terms of MSE.

The present paper provides a first proof of concept of our framework, on a very simple dataset (MNIST) with a very simple VAE. More experiments are needed to:
\begin{itemize}
    \item Verify that the framework preserves its qualitative advantages on more high-dimensional datasets (like CelebA, Fashion MNIST, etc.), and a larger selection of inverse problems.
    \item Improve the quality of the prior model by using more elaborate variations of variational autoencoders which mix the VAE framework with normalizing flows \citep{TwoStageVAE}, adversarial training \citep{Pu2017,Pu2017a,PGA-Zhang2019}, or BiGANs \cite{Donahue2019}.
\end{itemize}
All variations of the VAE framework cited above have the potential to improve the quality of our generative model, and to reduce the gap between $J_1$ and $J_2$. In particular, the adversarially symmetric VAE \citep{Pu2017,Pu2017a} proves that when learning reaches convergence the autoencoder approximation is exact, meaning that Assumption~\ref{z-convexity} would become true.

When compared to other decoupled plug \& play approaches that solve inverse problems using NN-based priors, our approach is constrained in different ways:\\
\emph{(a)} In a certain sense our approach is \emph{less constrained} than existing decoupled approaches since we do not require to retrain the NN-based denoiser to enforce any particular property to ensure convergence: \citet{ryu2019plug} requires the denoiser's residual operator to be non-expansive, and \citet{gupta2018cnn,shah2018solving} require the denoiser to act as a projector. The effect of these modifications to the denoiser on the quality of the underlying image prior has never been studied in detail and chances are that such constraints degrade it. Our method only requires a variational autoencoder without any further constraints, and the quality and expressiveness of this prior can be easily checked by sampling and reconstruction experiments. Checking the quality of the prior is a much more difficult task for \citet{ryu2019plug,gupta2018cnn,shah2018solving} which rely on an implicit prior, and do not provide a generative model.\\
\emph{(b)} On the other hand our method is \emph{more constrained} in the sense that it relies on a generative model of a \emph{fixed size}. Even if the generator and encoder are both convolutional neural networks, training and testing the same model on images of different sizes is a priori not possible because the latent space has a fixed dimension and a fixed distribution. As a future work we plan to explore different ways to address this limitation. The most straightforward way is to use our model to learn a prior of image patches of a fixed size and stitch this model via aggregation schemes like in EPLL~\cite{Zoran2011} to obtain a global prior model for images of any size. Alternatively we can use hierarchical generative models like in \cite{Karras2018} or resizable ones like in \cite{Bergmann2017}, and adapt our framework accordingly.

%% file: newdemo.tex
\section{Convergence Proofs}\label{sec:proofs}

\textcite{Gorski2007} establishes a general result on the simple alternate convex search (ACS) algorithm which consist in iterating the wollowing two steps:
\begin{itemize}
\item \emph{$\vx$-minimization} \emph{i.e.} $\vx_{n+1} = \argmin_{\vx} J_{1}(\vx,\vz_{n})$ (\emph{i.e.} solve problem~\eqref{eq:xmin}) 
\item \emph{$\vz$-minimization} \emph{i.e.} $\vz_{n+1} = \argmin_{\vz} J_{1}(\vx_{n+1},\vz)$ (\emph{i.e.} solve problem~\eqref{eq:zmin-exact})
\end{itemize}
until convergence for any continuous functional $J_1$ which is either bi-convex or which allows to solve both partial minimization subproblems exactly.

Under certain conditions, Algorithms 1, 2, and 3 are actually implementations of the ACS algorithm, and the general result stated in \parencite[Theorem 4.9]{Gorski2007} holds.

In the sequel we state the general theorem and then we verify the validity of the different hypotheses.

\begin{theorem}[Convergence of ACS]\label{thm:Gorski2007}
Let $X\subseteq\R^{n}$ and $Z\subseteq\R^{m}$ be closed sets and $J_{1}:X\times Z \to \R$ be continuous.
Let the optimization problems~\eqref{eq:xmin}~and~\eqref{eq:zmin-exact} be solvable.
\begin{enumerate}
\item If the sequence $\lbrace (\vx_{n},\vz_{n}) \rbrace_{n\in\NN}$ generated by ACS is contained in a compact set then the sequence has at least one accumulation point.
\item In addition suppose that for each accumulation point $(\vx^{*},\vz^{*})$:\\
either the optimal solution of~\eqref{eq:xmin} with $\vz=\vz^{*}$ is unique\\
or the optimal solution of~\eqref{eq:zmin-exact} with $\vx=\vx^{*}$ is unique.\\
Then all accumulation points are partial optima and have the same function value.
\item Furthermore if \\
\emph{(i)} $J_1$ is differentiable and bi-convex,  
and \\
\emph{(ii)} for each accumulation point  $(\vx^{*},\vz^{*})$ the optimal solutions of \emph{both}~\eqref{eq:xmin} with $\vz=\vz^{*}$ \emph{and}~\eqref{eq:zmin-exact} with $\vx=\vx^{*}$ are unique; \\
then:\\
\emph{(a)} the set of accumulation points is a connected compact set, and \\
\emph{(b)} all accumulation points in the interior of $X \times Z$ are stationary points

\end{enumerate}
\end{theorem}

\begin{proof}
This is the central result in \citep{Gorski2007}, proven in Theorem 4.9 and corollary 4.10.
\end{proof}

In the sequel we adopt the common assumption that all neural networks used in this work are composed of a finite number $d$ of layers, each layer being composed of: 
\emph{(a)} a linear operator (\emph{e.g.} convolutional or fully connected layer), followed by
\emph{(b)} a non-linear $L$-Lipschitz component-wise activation function with $0<L<\infty$.

Therefore we have the following property:

\begin{property}\label{lipschitz-nn}
For any neural network $f_\theta$ with parameters $\theta$ having the structure described above: \\
There exists a constant $C_\theta$ such that $\forall \vu$,
$$ \|f_\theta(\vu) \|_2 \leq C_\theta \| \vu \|_2.$$
\end{property}

Concerning activation functions we use two kinds:
\begin{itemize}
    \item continously differentiable activations like ELU, or
    \item continuous but non-differentiable activations like ReLU
\end{itemize}

\medskip 

\begin{lemma}\label{x-convexity}
$J_1$ is continuous. In addition
$\vx \mapsto J_1(\vx,\vz)$ is 
convex for all $\vz$,
and the map $\vx_{\min} : \vz \mapsto \argmin_\vx J_1(\vx,\vz)$ is single-valued. 
In addition for continuously differentiable activation functions, $J_1$ is continuously differentiable.
\end{lemma}
\begin{proof}
By construction $J_1$ is a composition of neural networks (which are composed of linear operators and continuous activations), linear and quadratic operations. Hence $J_1$ is continuous with respect to both variables, and continuously differentiable if the activation functions are so.
In addition $J_1$ is quadratic in $\vx$ for any fixed $\vz$.
The closed form in equation~\eqref{eq:xmin} shows that the mapping $\vx_{\min}$ is single-valued. It is also continuous because $\muDecoder$, $\SigmaDecoder$ are continuous neural networks.

\end{proof}

\begin{lemma}\label{z-convexity}
$J_2$ is continuous. In addition
$\vz \mapsto J_2(\vx,\vz)$ is convex for all $\vx$, and the map $\vz_{\min} : \vx \mapsto \argmin_\vz J_2(\vx,\vz)$ is single-valued. 
\end{lemma}
\begin{proof}
By construction $J_2$ is a composition of neural networks (which are composed of linear operators and continuous activations), linear and quadratic operations. Hence $J_1$ is continuous with respect to both variables.
In addition $J_2$ is quadratic in $\vz$ for any fixed $\vx$.
The closed form in equation~\eqref{eq:zmin} shows that the mapping is single-valued. It is also continuous because $\muEncoder$, $\SigmaEncoder$ are CNNs composed of convolutions and ReLUs, which are continuous functions.
\end{proof}

\medskip

\begin{lemma}\label{coercivity}
$J_1(\vx,\vz)$ is coercive.
\end{lemma}
\begin{proof}
If it was not coercive, then we could find a sequence $(\vx_{k},\vz_{k}) \to \infty$ such that $J_1(\vx_{k},\vz_{k})$ is bounded.
As a consequence all three (non-negative) terms are bounded. In particular $\|\vz_{k}\|$ is bounded, which means that $\vx_{k} \to \infty$.

From Property~\ref{lipschitz-nn}, $\muDecoder(\vz_k)$ and $\SigmaDecoder(\vz_k)$ are bounded for bounded $\vz_k$.

Now since $\muDecoder(\vz_k)$ and $\SigmaDecoder(\vz_k)$ are bounded and $\vx_{k} \to \infty$, we conclude that $\Hcoupling_{\decoderParams}(\vx_{k},\vz_{k}) \to \infty$ which contradicts our initial assumption.

As a consequence $J_1$ is coercive.
\end{proof}

\medskip

\begin{lemma}[Monotonicity]\label{monotonicity}  The sequence generated by Algorithms~\ref{alg:JPMAP2} and \ref{alg:JPMAP3} is non-increasing.
Under Assumption~\ref{exact-approximation}, Algorithm~\ref{alg:JPMAPexact} is also non-increasing.
\end{lemma}

\begin{proof}
Under assumption~\ref{exact-approximation}, Algorithm~\ref{alg:JPMAPexact} is exactly the ACS algorithm which obviously ensures monotonicity.

Step 6 in Algorithm~\ref{alg:JPMAP2} obviously makes the energy decrease. So does step 5 because
$$J_{1}(\vx_{n},\vz_{n+1}) \leq J_{1}(\vx_{n},\vz^{0}) \leq J_{1}(\vx_{n},\vz_{n}) $$

Step 10 in Algorithm~\ref{alg:JPMAP3} ensures that the $\vx$-update decreases the energy.
Step 5  in Algorithm~\ref{alg:JPMAP3} ensures that the $\vz$-update decreases the energy.
\end{proof}

\begin{proof}[\textbf{Proof of proposition~\ref{thm:convergence-exact}}] \hfill \break

\emph{1. Compact domain and accumulation points }

Theorem~\ref{thm:Gorski2007} applies to closed subsets $X$ and $Z$ whereas our algorithm does not restrict the domain.

This is not a big issue because the monotonicity and coercivity allow to show that the sequence is actually bounded.

Indeed, by definition of the coercivity (Lemma~\ref{coercivity}), the level set $S = \{(x,z)\mid J(x,z)\leq J_1(x_0,z_0)\}$ is bounded. The monotonicity of the sequence $(J_1(x_k,z_k))$ (Lemma~\ref{monotonicity}) implies that $(x_k,z_k)\in S$ for any $k$. Moreover, $S$ is closed, thus compact, since $J_1$ is continuous. The existence of an accumulation point is straight-forward.

This proves the first part in Proposition~\ref{thm:convergence-exact}.

\emph{2. Properties of the set of accumulation points.}

Since the sequence $\left\lbrace (\vx_{n},\vz_{n})  \right\rbrace$ is bounded, there exist compact sets $X_{0}$ and $Z_{0}$ which contain the entire sequence. Now  consider the dilated compact sets
$$ X = \lbrace \vx\,:\, \exists \ve\in\RR^{n},\, \|\ve\|\leq \epsilon, \text{and}\, (\vx+\ve) \in X_{0} \rbrace $$
$$ Z = \lbrace \vz\,:\, \exists \ve\in\RR^{m},\, \|\ve\|\leq \epsilon, \text{and}\, (\vz+\ve) \in Z_{0} \rbrace.$$
Then the sequence $\left\lbrace (\vx_{n},\vz_{n})  \right\rbrace$ and all its accumulation points are all contained in the interior of $X \times Z$.

Now we can apply Theorem~\ref{thm:Gorski2007} to $J_{1}$ on the restricted domain $X\times Z$.
Indeed, from Assumption~\ref{exact-approximation} we know that $J_{1}=J_{2}$, which means that Algorithm~\ref{alg:JPMAPexact} is an implementation of ACS.
In addition Lemmas~\ref{x-convexity}~and~\ref{z-convexity} show that  optimization subproblems~\eqref{eq:xmin}~and~\eqref{eq:zmin-exact} are solvable and have unique solutions.

Therefore all hypotheses of Theorem~\ref{thm:Gorski2007} are met, which shows the second part of Proposition~\ref{thm:convergence-exact}.
\end{proof}

\medskip

\begin{proof}[\textbf{Proof of proposition~\ref{thm:convergence-approx}}] \hfill \break

\emph{1. Compact domain and accumulation points }

First note that Algorithms 2 and 3 (for $n>n_{\min}$) are particular implementations of ACS on $J_{1}$.  
Therefore the monotonicity of the sequence  $J_{1}(\vx_{n},\vz_{n})$ (Lemma~\ref{monotonicity}) and the coercivity of $J_{1}$ are sufficient to show the first part of proposition~\ref{thm:convergence-approx} exactly like in proposition~\ref{thm:convergence-exact}.
This also shows that the sequence $\left\lbrace (\vx_{n},\vz_{n})  \right\rbrace$ is bounded.

\emph{2. Accumulation points are partial optima}

Since the sequence $\left\lbrace (\vx_{n},\vz_{n})  \right\rbrace$ is bounded, there exist compact sets $X_{0}$ and $Z_{0}$ which contain the entire sequence. Now  consider the dilated compact sets
$$ X = \lbrace \vx\,:\, \exists \ve\in\RR^{n},\, \|\ve\|\leq \epsilon, \text{and}\, (\vx+\ve) \in X_{0} \rbrace $$
$$ Z = \lbrace \vz\,:\, \exists \ve\in\RR^{m},\, \|\ve\|\leq \epsilon, \text{and}\, (\vz+\ve) \in Z_{0} \rbrace.$$
Then the sequence $\left\lbrace (\vx_{n},\vz_{n})  \right\rbrace$ and all its accumulation points are all contained in the interior of $X \times Z$.

Now we can apply parts 2 and 3 of Theorem~\ref{thm:Gorski2007} to $J_{1}$ on the restricted domain $X\times Z$.

Indeed, from Lemma~\ref{x-convexity}, the $\vx$-minimization subproblem~\eqref{eq:xmin} is solvable with unique solution.
In addition from Assumption~\ref{z-convexity1}(A), the  $\vz$-minimization subproblem~\eqref{eq:zmin-exact} is solvable and algorithms 2 and 3 provide that solution for each fixed $\vx$.

Therefore all hypotheses for part 2  of Theorem~\ref{thm:Gorski2007} are met, which shows that all accumulation points are partial optima, and they all have the same function value.

\emph{3. Accumulation points are stationary points.}

Known facts:
\begin{enumerate}
    \item $J_1$ differentiable: $\nabla J_1(x_k,z_k) = \left(\frac{\partial J_1}{\partial x}(x_k,z_k),\frac{\partial J_1}{\partial z}(x_k,z_k)\right)$
    \item $x\mapsto J_1(x,z)$ convex:
    \[\forall\, y, J_1(y,z)\geq J_1(x,z) + \left\langle \frac{\partial J_1}{\partial x}(x,z),y-x\right\rangle\]
    Hence a minimizer is a stationary point.
    \item Unicity of partial minimizers: in particular, thanks to  the convexity of $J_1$ -- stationary points are minimizers
\[
    \frac{\partial J_1}{\partial x}(x^*,\hat z) = 0 \qquad\Longrightarrow\qquad \frac{\partial J_1}{\partial x}(x,\hat z) \neq 0\quad
    \text{if }x\neq x^*
    \]
    \item $J_1$ coercive and descent scheme: $(x_k,z_k)_k$ and $(x_{k+1},z_k)_k$ are bounded thus have convergent subsequences
    \item Decrease property:
    \[J_1(x_k,z_k) \geq J_1(x_{k+1},z_k) \geq J_1(x_{k+1},z_{k+1})
    \]
    If $J_1$ is lowerbounded,
    \[
    \lim_{k\to+\infty} 
    J_1(x_k,z_k) =     \lim_{k\to+\infty}  J_1(x_{k+1},z_k) =     \lim_{k\to+\infty}  J_1(x_{k+1},z_{k+1})
    \]
\item Fermat's rule: 
\[
\frac{\partial J_1}{\partial x}(x_{k+1},z_k) = 0 \quad\text{ and }\quad
    \frac{\partial J_1}{\partial z}(x_{k+1},z_{k+1}) = 0 \]
\end{enumerate}
\begin{itemize}\item Let $(x^*,z^*)$ be an accumulation point of $(x_{k},z_{k})_k$. If $(x_{k_j+1},z_{k_j+1})_j$ converges to $(x^*,z^*)$,  $(x_{k_j})_j$ and $(z_{k_j})_j$ are bounded thus have convergent subsequences. So there exists a sequence $\{j_n\}_n$ such that (double extraction of subsequence)
    \[
    \lim_{n\to+\infty} x_{k_{j_n}+1} = x^*
    \quad\text{ and }\quad
    \lim_{n\to+\infty} x_{k_{j_n}} = \hat x\]
    \[
    \lim_{n\to+\infty} z_{k_{j_n}+1} = z^*
    \quad\text{ and }\quad
    \lim_{n\to+\infty} z_{k_{j_n}} = \hat z\]
According to Fact 5, $J_1(x^*,z^*) = J_1(x^*,\hat z) = J_1(\hat x,\hat z)$.

\item Partial minimizers:
\[\forall\,x, J_1(x,z_{k_{j_n}}) \geq J_1(x_{k_{j_n}+1},z_{k_{j_n}}) \]
By taking the limit (since $J_1$ is continuous)
\[\forall\,x, J_1(x,\hat z) \geq J_1(x^*,\hat z) \]
Thus, $x^*$ is the unique minimizer of $J_1(\cdot, \hat z)$. By unicity of the partial minimizer, $\hat x = x^*$.
\[\forall\,z, J_1(x_{k_{j_n}+1},z) \geq J_1(x_{k_{j_n}+1},z_{k_{j_n}+1}) \]
By taking the limit (since $J_1$ is continuous)
\[\forall\,z, J_1(x^*,z) \geq J_1(x^*,z^*) \]
Thus, $z^*$ is the unique minimizer of $J_1(x^*,\cdot)$. By unicity of the partial minimizer (Assumption \ref{z-convexity1}(B)), $\hat z = z^*$.
\item Since $J_1(\cdot,z^*)$ is convex,
$x^*$ is the minimizer of $J_1(\cdot,z^*)$ iff $\frac{\partial J_1}{\partial x}(x^*,z^*) = 0$
\item If $z\mapsto J_1(x,z)$ is continuously differentiable, then 
\[\lim_{n\to +\infty} \frac{\partial J_1}{\partial z}(x_{k_{j_n}+1},z_{k_{j_n}+1}) = \frac{\partial J_1}{\partial z}(x^*,z^*) = 0\]
\item As a consequence, 
\[\nabla J_1(x^*,z^*) = 0\]
Otherwise said,
\begin{quote} 
If $J_1(\cdot, z)$ is convex, $J_1$ is continuouly differentiable and coercive, and if the partial minimizers are all unique, then 
any accumulation point of the sequence $(x_k,z_k)_k$ is a stationary point.
\end{quote}
\end{itemize}
\end{proof}

%% file: main.bbl
\begin{thebibliography}{39}
\providecommand{\natexlab}[1]{#1}
\providecommand{\url}[1]{\texttt{#1}}
\expandafter\ifx\csname urlstyle\endcsname\relax
  \providecommand{\doi}[1]{doi: #1}\else
  \providecommand{\doi}{doi: \begingroup \urlstyle{rm}\Url}\fi

\bibitem[Aguerrebere et~al.(2017)Aguerrebere, Almansa, Delon, Gousseau, and
  Muse]{Aguerrebere2014b}
Cecilia Aguerrebere, Andres Almansa, Julie Delon, Yann Gousseau, and Pablo
  Muse.
\newblock {A Bayesian Hyperprior Approach for Joint Image Denoising and
  Interpolation, With an Application to HDR Imaging}.
\newblock \emph{IEEE Transactions on Computational Imaging}, 3\penalty0
  (4):\penalty0 633--646, dec 2017.
\newblock ISSN 2333-9403.
\newblock \doi{10.1109/TCI.2017.2704439}.
\newblock URL \url{https://nounsse.github.io/HBE_project/}.

\bibitem[Bergmann et~al.(2017)Bergmann, Jetchev, and Vollgraf]{Bergmann2017}
Urs Bergmann, Nikolay Jetchev, and Roland Vollgraf.
\newblock {Learning Texture Manifolds with the Periodic Spatial GAN}.
\newblock \emph{(ICML) International Conference on Machine Learning},
  1:\penalty0 722--730, may 2017.

\bibitem[Bigdeli \& Zwicker(2017)Bigdeli and Zwicker]{Bigdeli2017a}
Siavash~Arjomand Bigdeli and Matthias Zwicker.
\newblock {Image Restoration using Autoencoding Priors}.
\newblock Technical report, 2017.

\bibitem[Bigdeli et~al.(2017)Bigdeli, Jin, Favaro, and Zwicker]{Bigdeli2017}
Siavash~Arjomand Bigdeli, Meiguang Jin, Paolo Favaro, and Matthias Zwicker.
\newblock {Deep Mean-Shift Priors for Image Restoration}.
\newblock In \emph{(NIPS) Advances in Neural Information Processing Systems
  30}, pp.\  763--772, sep 2017.
\newblock URL
  \url{http://papers.nips.cc/paper/6678-deep-mean-shift-priors-for-image-restoration}.

\bibitem[Bora et~al.(2017)Bora, Jalal, Price, and Dimakis]{bora2017compressed}
Ashish Bora, Ajil Jalal, Eric Price, and Alexandros~G Dimakis.
\newblock Compressed sensing using generative models.
\newblock In \emph{(ICML) International Conference on Machine Learning},
  volume~2, pp.\  537--546. JMLR. org, 2017.
\newblock ISBN 9781510855144.

\bibitem[Chambolle(2004)]{Chambolle04}
A~Chambolle.
\newblock {An algorithm for total variation minimization and applications}.
\newblock \emph{Journal of Mathematical Imaging and Vision}, 20:\penalty0
  89--97, 2004.
\newblock \doi{10.1023/B:JMIV.0000011325.36760.1e}.

\bibitem[{Chan} et~al.(2017){Chan}, {Wang}, and {Elgendy}]{chan2017plug}
S.~H. {Chan}, X.~{Wang}, and O.~A. {Elgendy}.
\newblock Plug-and-play admm for image restoration: Fixed-point convergence and
  applications.
\newblock \emph{IEEE Transactions on Computational Imaging}, 3\penalty0
  (1):\penalty0 84--98, March 2017.
\newblock ISSN 2333-9403.
\newblock \doi{10.1109/TCI.2016.2629286}.

\bibitem[Chen \& Pock(2017)Chen and Pock]{Chen2017}
Yunjin Chen and Thomas Pock.
\newblock {Trainable Nonlinear Reaction Diffusion: A Flexible Framework for
  Fast and Effective Image Restoration}.
\newblock \emph{IEEE Transactions on Pattern Analysis and Machine
  Intelligence}, 39\penalty0 (6):\penalty0 1256--1272, 2017.
\newblock ISSN 01628828.
\newblock \doi{10.1109/TPAMI.2016.2596743}.

\bibitem[Clevert et~al.(2016)Clevert, Unterthiner, and Hochreiter]{Clevert2016}
Djork-Arn{\'{e}} Clevert, Thomas Unterthiner, and Sepp Hochreiter.
\newblock {Fast and Accurate Deep Network Learning by Exponential Linear Units
  (ELUs)}.
\newblock In \emph{(ICLR) International Conference on Learning
  Representations}, nov 2016.

\bibitem[Dai \& Wipf(2019)Dai and Wipf]{TwoStageVAE}
Bin Dai and David Wipf.
\newblock {Diagnosing and Enhancing VAE Models}.
\newblock \emph{ICLR}, pp.\  1--42, 2019.
\newblock URL \url{https://openreview.net/forum?id=B1e0X3C9tQ}.

\bibitem[Diamond et~al.(2017)Diamond, Sitzmann, Heide, and
  Wetzstein]{diamond2017unrolled}
Steven Diamond, Vincent Sitzmann, Felix Heide, and Gordon Wetzstein.
\newblock Unrolled optimization with deep priors.
\newblock 2017.

\bibitem[Donahue \& Simonyan(2019)Donahue and Simonyan]{Donahue2019}
Jeff Donahue and Karen Simonyan.
\newblock {Large Scale Adversarial Representation Learning}.
\newblock 2019.
\newblock URL \url{http://arxiv.org/abs/1907.02544}.

\bibitem[Dong et~al.(2014)Dong, Loy, He, and Tang]{dong2014learning}
Chao Dong, Chen~Change Loy, Kaiming He, and Xiaoou Tang.
\newblock Learning a deep convolutional network for image super-resolution.
\newblock In \emph{European conference on computer vision}, pp.\  184--199.
  Springer, 2014.

\bibitem[Gao et~al.(2019)Gao, Tao, Shen, and Jia]{gao2019dynamic}
Hongyun Gao, Xin Tao, Xiaoyong Shen, and Jiaya Jia.
\newblock Dynamic scene deblurring with parameter selective sharing and nested
  skip connections.
\newblock In \emph{Proceedings of the IEEE Conference on Computer Vision and
  Pattern Recognition}, pp.\  3848--3856, 2019.

\bibitem[Gharbi et~al.(2016)Gharbi, Chaurasia, Paris, and
  Durand]{gharbi2016deep}
Micha{\"e}l Gharbi, Gaurav Chaurasia, Sylvain Paris, and Fr{\'e}do Durand.
\newblock Deep joint demosaicking and denoising.
\newblock \emph{ACM Transactions on Graphics (TOG)}, 35\penalty0 (6):\penalty0
  191, 2016.

\bibitem[Gilton et~al.(2019)Gilton, Ongie, and Willett]{gilton2019neumann}
Davis Gilton, Greg Ongie, and Rebecca Willett.
\newblock Neumann networks for inverse problems in imaging.
\newblock 2019.

\bibitem[Gorski et~al.(2007)Gorski, Pfeuffer, and Klamroth]{Gorski2007}
Jochen Gorski, Frank Pfeuffer, and Kathrin Klamroth.
\newblock {Biconvex sets and optimization with biconvex functions: a survey and
  extensions}.
\newblock \emph{Mathematical Methods of Operations Research}, 66\penalty0
  (3):\penalty0 373--407, nov 2007.
\newblock ISSN 1432-2994.
\newblock \doi{10.1007/s00186-007-0161-1}.

\bibitem[Gregor \& LeCun(2010)Gregor and LeCun]{gregor2010learning}
Karol Gregor and Yann LeCun.
\newblock Learning fast approximations of sparse coding.
\newblock In \emph{Proceedings of the 27th International Conference on
  International Conference on Machine Learning}, pp.\  399--406. Omnipress,
  2010.

\bibitem[Gupta et~al.(2018)Gupta, Jin, Nguyen, McCann, and Unser]{gupta2018cnn}
Harshit Gupta, Kyong~Hwan Jin, Ha~Q Nguyen, Michael~T McCann, and Michael
  Unser.
\newblock Cnn-based projected gradient descent for consistent ct image
  reconstruction.
\newblock \emph{IEEE transactions on medical imaging}, 37\penalty0
  (6):\penalty0 1440--1453, 2018.
\newblock \doi{10.1109/TMI.2018.2832656}.

\bibitem[Karras et~al.(2017)Karras, Aila, Laine, and Lehtinen]{Karras2018}
Tero Karras, Timo Aila, Samuli Laine, and Jaakko Lehtinen.
\newblock {Progressive Growing of GANs for Improved Quality, Stability, and
  Variation}.
\newblock \emph{(ICLR) International Conference on Learning Representations},
  10\penalty0 (2):\penalty0 327--331, oct 2017.
\newblock URL \url{https://openreview.net/forum?id=Hk99zCeAb}.

\bibitem[Kingma \& Welling(2013)Kingma and Welling]{Kingma2014}
Diederik~P Kingma and Max Welling.
\newblock {Auto-Encoding Variational Bayes}.
\newblock In \emph{(ICLR) International Conference on Learning
  Representations}, number~Ml, pp.\  1--14, dec 2013.
\newblock ISBN 1312.6114v10.
\newblock \doi{10.1051/0004-6361/201527329}.

\bibitem[Krizhevsky et~al.(2012)Krizhevsky, Sutskever, and
  Hinton.]{Krizhevsky2012}
Alex Krizhevsky, Ilya Sutskever, and Geoffrey~E. Hinton.
\newblock {Imagenet classification with deep convolutional neural networks}.
\newblock \emph{(NIPS) Advances in neural information processing systems}, pp.\
   1097--1105, 2012.
\newblock ISSN 10495258.

\bibitem[Lecun et~al.(1998)Lecun, Bottou, Bengio, and Haffner]{MNIST}
Yann Lecun, Leon Bottou, Yoshua Bengio, and P.~Haffner.
\newblock {Gradient-based learning applied to document recognition}.
\newblock \emph{Proceedings of the IEEE}, 86\penalty0 (11):\penalty0
  2278--2324, 1998.
\newblock ISSN 00189219.
\newblock \doi{10.1109/5.726791}.

\bibitem[Louchet \& Moisan(2013)Louchet and Moisan]{Louchet2013}
C{\'{e}}cile Louchet and Lionel Moisan.
\newblock {Posterior expectation of the total variation model: Properties and
  experiments}.
\newblock \emph{SIAM Journal on Imaging Sciences}, 6\penalty0 (4):\penalty0
  2640--2684, dec 2013.
\newblock ISSN 19364954.
\newblock \doi{10.1137/120902276}.

\bibitem[Meinhardt et~al.(2017)Meinhardt, Moller, Hazirbas, and
  Cremers]{meinhardt2017learning}
Tim Meinhardt, Michael Moller, Caner Hazirbas, and Daniel Cremers.
\newblock Learning proximal operators: Using denoising networks for
  regularizing inverse imaging problems.
\newblock In \emph{(ICCV) International Conference on Computer Vision}, pp.\
  1781--1790, 2017.
\newblock \doi{10.1109/ICCV.2017.198}.
\newblock URL
  \url{http://openaccess.thecvf.com/content_iccv_2017/html/Meinhardt_Learning_Proximal_Operators_ICCV_2017_paper.html}.

\bibitem[Pereyra(2016)]{Pereyra2016}
Marcelo Pereyra.
\newblock {Proximal Markov chain Monte Carlo algorithms}.
\newblock \emph{Statistics and Computing}, 26\penalty0 (4):\penalty0 745--760,
  jul 2016.
\newblock ISSN 0960-3174.
\newblock \doi{10.1007/s11222-015-9567-4}.
\newblock URL \url{http://dx.doi.org/10.1007/s11222-015-9567-4}.

\bibitem[Pu et~al.(2017{\natexlab{a}})Pu, Wang, Henao, Chen, Gan, Li, and
  Carin]{Pu2017}
Yunchen Pu, Weiyao Wang, Ricardo Henao, Liqun Chen, Zhe Gan, Chunyuan Li, and
  Lawrence Carin.
\newblock {Adversarial symmetric variational autoencoder}.
\newblock In \emph{(NIPS) Advances in Neural Information Processing Systems},
  pp.\  4331--4340, 2017{\natexlab{a}}.

\bibitem[Pu et~al.(2017{\natexlab{b}})Pu, Wang, Henao, Chen, Gan, Li, and
  Carin]{Pu2017a}
Yunchen Pu, Weiyao Wang, Ricardo Henao, Liqun Chen, Zhe Gan, Chunyuan Li, and
  Lawrence Carin.
\newblock {Adversarial symmetric variational autoencoder}.
\newblock In \emph{(NIPS) Advances in Neural Information Processing Systems},
  volume 2017-Decem, pp.\  4331--4340, 2017{\natexlab{b}}.

\bibitem[Reehorst \& Schniter(2018)Reehorst and
  Schniter]{reehorst2018regularization}
Edward~T Reehorst and Philip Schniter.
\newblock Regularization by denoising: Clarifications and new interpretations.
\newblock \emph{IEEE Transactions on Computational Imaging}, 5\penalty0
  (1):\penalty0 52--67, 2018.
\newblock \doi{10.1109/TCI.2018.2880326}.

\bibitem[Rudin et~al.(1992)Rudin, Osher, and Fatemi]{Rudin1992}
Leonid~I. Rudin, Stanley Osher, and Emad Fatemi.
\newblock {Nonlinear total variation based noise removal algorithms}.
\newblock \emph{Physica D: Nonlinear Phenomena}, 60\penalty0 (1-4):\penalty0
  259--268, 1992.
\newblock ISSN 01672789.
\newblock \doi{10.1016/0167-2789(92)90242-F}.

\bibitem[Ryu et~al.(2019)Ryu, Liu, Wang, Chen, Wang, and Yin]{ryu2019plug}
Ernest~K. Ryu, Jialin Liu, Sicheng Wang, Xiaohan Chen, Zhangyang Wang, and
  Wotao Yin.
\newblock Plug-and-play methods provably converge with properly trained
  denoisers.
\newblock In \emph{Proceedings of the 36th International Conference on Machine
  Learning, {ICML} 2019, 9-15 June 2019, Long Beach, California, {USA}}, pp.\
  5546--5557, 2019.
\newblock URL \url{http://proceedings.mlr.press/v97/ryu19a.html}.

\bibitem[Schwartz et~al.(2018)Schwartz, Giryes, and
  Bronstein]{schwartz2018deepisp}
Eli Schwartz, Raja Giryes, and Alex~M Bronstein.
\newblock Deepisp: Toward learning an end-to-end image processing pipeline.
\newblock \emph{IEEE Transactions on Image Processing}, 28\penalty0
  (2):\penalty0 912--923, 2018.

\bibitem[Shah \& Hegde(2018)Shah and Hegde]{shah2018solving}
Viraj Shah and Chinmay Hegde.
\newblock Solving linear inverse problems using gan priors: An algorithm with
  provable guarantees.
\newblock In \emph{2018 IEEE International Conference on Acoustics, Speech and
  Signal Processing (ICASSP)}, pp.\  4609--4613. IEEE, 2018.

\bibitem[Teodoro et~al.(2018)Teodoro, Bioucas-Dias, and
  Figueiredo]{Teodoro2018scene}
Afonso~M. Teodoro, Jos{\'{e}}~M. Bioucas-Dias, and M{\'{a}}rio A.~T.
  Figueiredo.
\newblock {Scene-Adapted Plug-and-Play Algorithm with Guaranteed Convergence:
  Applications to Data Fusion in Imaging}, jan 2018.

\bibitem[Zhang et~al.(2017{\natexlab{a}})Zhang, Zuo, Chen, Meng, and
  Zhang]{zhang2017beyond}
Kai Zhang, Wangmeng Zuo, Yunjin Chen, Deyu Meng, and Lei Zhang.
\newblock Beyond a gaussian denoiser: Residual learning of deep cnn for image
  denoising.
\newblock \emph{IEEE Transactions on Image Processing}, 26\penalty0
  (7):\penalty0 3142--3155, 2017{\natexlab{a}}.

\bibitem[Zhang et~al.(2017{\natexlab{b}})Zhang, Zuo, Gu, and Zhang]{Zhang2017}
Kai Zhang, Wangmeng Zuo, Shuhang Gu, and Lei Zhang.
\newblock {Learning Deep CNN Denoiser Prior for Image Restoration}.
\newblock In \emph{(CVPR) IEEE Conference on Computer Vision and Pattern
  Recognition}, pp.\  2808--2817. IEEE, apr 2017{\natexlab{b}}.
\newblock ISBN 978-1-5386-0457-1.
\newblock \doi{10.1109/CVPR.2017.300}.
\newblock URL
  \url{http://openaccess.thecvf.com/content_cvpr_2017/html/Zhang_Learning_Deep_CNN_CVPR_2017_paper.html}.

\bibitem[Zhang et~al.(2018)Zhang, Zuo, and Zhang]{zhang2018ffdnet}
Kai Zhang, Wangmeng Zuo, and Lei Zhang.
\newblock Ffdnet: Toward a fast and flexible solution for cnn-based image
  denoising.
\newblock \emph{IEEE Transactions on Image Processing}, 27\penalty0
  (9):\penalty0 4608--4622, 2018.

\bibitem[Zhang et~al.(2019)Zhang, Zhang, Li, Bengio, and Paull]{PGA-Zhang2019}
Zijun Zhang, Ruixiang Zhang, Zongpeng Li, Yoshua Bengio, and Liam Paull.
\newblock {Perceptual Generative Autoencoders}.
\newblock In \emph{(ICLR) International Conference on Learning
  Representations}, pp.\  1--7, jun 2019.
\newblock URL \url{https://github.com/zj10/PGA}.

\bibitem[Zoran \& Weiss(2011)Zoran and Weiss]{Zoran2011}
Daniel Zoran and Yair Weiss.
\newblock {From learning models of natural image patches to whole image
  restoration}.
\newblock In \emph{2011 International Conference on Computer Vision}, pp.\
  479--486. IEEE, nov 2011.
\newblock ISBN 978-1-4577-1102-2.
\newblock \doi{10.1109/ICCV.2011.6126278}.
\newblock URL
  \url{http://people.csail.mit.edu/danielzoran/EPLLICCVCameraReady.pdf}.

\end{thebibliography}
